\documentclass[11pt,leqno,oneside]{amsart}
\usepackage{graphicx}
\usepackage{indentfirst,csquotes}

\usepackage{amssymb,amsthm,amsmath}
\usepackage{paralist,hyperref,titlesec,fancyhdr,etoolbox}

\usepackage{titlesec}

\titleformat{\section}[hang]
  {\normalfont\LARGE\bfseries}{\thesection}{1em}{}

\titleformat{\subsection}[hang]
  {\normalfont\Large\bfseries}{\thesubsection}{1em}{}

\titleformat{\subsubsection}[hang]
  {\normalfont\normalsize\bfseries}{\thesubsubsection}{1em}{}

\titlespacing*{\section}{0pt}{1.5ex plus 1ex minus .2ex}{1ex}
\titlespacing*{\subsection}{0pt}{1.25ex plus 1ex minus .2ex}{1ex}
\titlespacing*{\subsubsection}{0pt}{1ex plus .5ex minus .2ex}{1ex}

\hypersetup{ colorlinks=true, linkcolor=black, filecolor=black, urlcolor=black }

\usepackage{lipsum}
\usepackage{float}
\usepackage{setspace} 
\usepackage{amssymb}
\usepackage{amsmath}
\usepackage{booktabs}
\usepackage{amssymb,amsmath,amsfonts,amscd,verbatim}
\usepackage{url,upgreek}
\usepackage{graphicx}
\usepackage{mathrsfs}
\usepackage{amssymb}
\usepackage{amsmath}
\usepackage{bbm}
\usepackage{url,upgreek}
\usepackage{float}
\usepackage[font={small}]{caption}
\usepackage{geometry}
\usepackage{setspace}
\usepackage{amssymb,amsmath,amsfonts,amscd,verbatim}
\usepackage{url,upgreek}
\usepackage{amsmath,amsfonts,amsthm,amssymb}
\usepackage{setspace}
\usepackage{Tabbing}
\usepackage{fancyhdr}
\usepackage{lastpage}
\usepackage{extramarks}
\usepackage{chngpage}
\usepackage{soul,color}
\usepackage{graphicx,float,wrapfig}
\usepackage{amsfonts}
\usepackage{bbold}
\usepackage{soul}
\newcommand\textopenone{\leavevmode\hbox{\small 1\kern-3.3pt\normalsize 1}}

\usepackage{bm}
\usepackage{amssymb,amsmath,amsfonts,amscd,verbatim}
\usepackage{url,upgreek}
\usepackage{graphicx}
\usepackage{mathrsfs}
\usepackage{amssymb,amsmath,amsfonts,amscd,verbatim}
\usepackage{url,upgreek}
\usepackage{amsmath,amsfonts,amsthm,amssymb}
\usepackage{setspace}
\usepackage{Tabbing}
\usepackage{fancyhdr}
\usepackage{lastpage}
\usepackage{extramarks}
\usepackage{chngpage}
\usepackage{soul,color}
\usepackage{graphicx,float,wrapfig}
\usepackage{amsfonts}
\usepackage{bbold}
\usepackage{hyperref}
\usepackage{mathtools}
\usepackage{bbm}
\usepackage{url,upgreek}
\usepackage{float}
\usepackage[font={small}]{caption}
\usepackage{geometry}
\usepackage{listings}
\usepackage{setspace}
\usepackage[table,xcdraw,dvipsnames]{xcolor}
\usepackage{color,soul}
\usepackage[linesnumbered,ruled,lined]{algorithm2e}
\SetKwInput{KwInput}{Input}                % Set the Input
\SetKwInput{KwOutput}{Output}              % set the Output

\usepackage{tikz}

\usepackage{siunitx,booktabs,threeparttable,caption}

\usepackage{bbm}

\makeatletter
\def\@settitle{%
  \begin{center}%
    \baselineskip=18pt
    \normalfont\bfseries\fontsize{16}{20}\selectfont
    \@title
  \end{center}%
}
\makeatother

\begin{document}
\title{A Multi-head Attention Fusion Network for Industrial Prognostics under Discrete Operational Conditions}

\author{Yuqi Su$^1$}
\author{Xiaolei Fang$^2$}

\let\thefootnote\relax
%\footnotetext{This work has been submitted to the IEEE for possible publication. Copyright may be transferred without notice, after which this version may no longer be accessible.}
\footnotetext{\textsuperscript{1}Operations Research Graduate Program, North Carolina State University, Raleigh, NC, USA.  Email: ysu25@ncsu.edu}
\footnotetext{\textsuperscript{2}Edward P. Fitts Department of Industrial and Systems Engineering, North Carolina State University, Raleigh, NC, USA. Email: xfang8@ncsu.edu}

\begin{abstract}
Complex systems such as aircraft engines, turbines, and industrial machinery often operate under dynamically changing conditions. These varying operating conditions can substantially influence degradation behavior and make prognostic modeling more challenging, as accurate prediction requires explicit consideration of operational effects. To address this issue, this paper proposes a novel multi-head attention-based fusion neural network. The proposed framework explicitly models and integrates three signal components: (1) the monotonic degradation trend, which reflects the underlying deterioration of the system; (2) discrete operating states, identified through clustering and encoded into dense embeddings; and (3) residual random noise, which captures unexplained variation in sensor measurements. The core strength of the framework lies in its architecture, which combines BiLSTM networks with attention mechanisms to better capture complex temporal dependencies. The attention mechanism allows the model to adaptively weight different time steps and sensor signals, improving its ability to extract prognostically relevant information. In addition, a fusion module is designed to integrate the outputs from the degradation-trend branch and the operating-state embeddings, enabling the model to capture their interactions more effectively. The proposed method is validated using a dataset from the NASA repository, and the results demonstrate its effectiveness.
\end{abstract}
\maketitle

\bigskip

\section{Introduction}\label{section4.1}

Prognostics and Health Management (PHM) have emerged as indispensable methodologies in modern industrial practices, particularly for enhancing the reliability, safety, and operational efficiency of complex engineering systems. The capability of accurately predicting the Remaining Useful Life (RUL) of machinery and equipment has profound implications across numerous sectors, including aerospace, automotive, energy, and manufacturing. Precise RUL predictions facilitate optimal maintenance schedules, reduce unnecessary downtime, enhance system availability, and ultimately contribute to significant cost reductions and improved operational safety \cite{jardine2006review, lee2014prognostics, pecht2010modeling}.

Traditionally, approaches to prognostics can be broadly categorized into model-based and data-driven methods. Model-based methods rely heavily on detailed physical or mathematical models that represent degradation mechanisms, employing techniques such as particle filtering, Kalman filtering, and stochastic modeling \cite{daigle2016qualitative, thelen2022integrating}. While these approaches offer substantial interpretability, their effectiveness is significantly constrained by the complexity of degradation phenomena and the need for substantial domain knowledge and accurate physical models \cite{pecht2010modeling}. Such limitations hinder their applicability to diverse real-world scenarios where operational conditions frequently vary and explicit physical models might be unavailable or excessively simplified.

In contrast, data-driven methods have gained increasing attention due to their flexibility and powerful predictive capabilities, especially in scenarios where comprehensive physical models are challenging to formulate. With the exponential growth of sensor data acquisition and storage capabilities, machine learning and deep learning approaches, such as neural networks, ensemble methods, and deep recurrent architectures, have demonstrated remarkable effectiveness in modeling complex temporal patterns and degradation trajectories \cite{zhang2018long, wang2018hybrid, su2024deep}. Among these, recurrent neural networks (RNNs) and specifically Long Short-Term Memory (LSTM) networks have become predominant tools due to their inherent capability to capture sequential dependencies and effectively manage long-term temporal information \cite{hochreiter1997long, zhao2019deep, yang2020remaining}.

Nevertheless, despite these advancements, current data-driven approaches often overlook critical aspects of degradation under variable operational conditions. Real-world engineering systems are rarely operated under constant conditions, and the operational variability significantly influences the degradation behavior and subsequent sensor signals. The presence of distinct operational states leads to state-dependent behavior in degradation trajectories. Failure to explicitly model such state-dependent variations limits the predictive accuracy and interpretability of current prognostic models, potentially leading to overly optimistic or misleading predictions of system health and RUL \cite{zhou2024adaptive, cheng2022multi, ellefsen2019remaining, li2018remaining}.

Guided by domain knowledge, condition-monitoring signals can be viewed as the superposition of three effects: a monotonic and smooth degradation trend that reflects irreversible wear; state-dependent shifts induced by changes in operating conditions; and residual noise. Recent research has explored hybrid or ensemble approaches that combine operational-condition clustering (for example, KMeans) with machine learning or deep learning. However, these methods typically either predict RUL independently of state or incorporate condition information only implicitly, without enforcing this decomposition \cite{kim2021multitask, zhu2024contrastive, huang2019bidirectional}. They also tend to emphasize single-point RUL prediction while ignoring the forecasting of full future signal trajectories beyond the observation cutoff, even though maintenance planning benefits from visibility into the likely evolution of sensor signals. As a result, there remains a gap in effectively capturing and explicitly separating the underlying monotonic degradation process from condition-dependent fluctuations and random noise within a unified and interpretable modeling framework.

Motivated by these gaps, this study introduces a deep learning framework designed both to model and predict the remaining useful life of systems operating under varying states and to forecast future sensor trajectories. The primary contribution is a multi-headed neural network architecture that systematically decomposes each signal into three interpretable components:  (1) an underlying monotonic degradation trend capturing the inherent irreversible deterioration; (2) discrete operational condition states identified via clustering methods, representing distinct operating regimes; and (3) residual random noise that captures unexplained variability. This explicit decomposition facilitates significantly improved interpretability and prediction accuracy, making it highly suitable for practical PHM applications.

The proposed methodology integrates state-of-the-art deep learning techniques, including attention-based Bidirectional LSTM networks for extracting deep temporal features and advanced embedding methods for condition state representation. To enhance predictive performance, we introduce an innovative fusion mechanism combining degradation trends and state embeddings, explicitly guiding the forecast output to reflect realistic state-dependent behavior. Furthermore, we employ asymmetric loss functions to penalize late RUL predictions, explicitly addressing safety-critical concerns where optimistic predictions may lead to severe operational consequences.

%To validate the effectiveness of our proposed framework, comprehensive numerical studies are conducted on the widely utilized NASA C-MAPSS (Commercial Modular Aero-Propulsion System Simulation) dataset, specifically the FD002 dataset known for its operational condition variability and complexity \cite{saxena2008damage}. Detailed analyses demonstrate not only significant improvements in predictive accuracy compared to conventional methods but also clear evidence of accurately capturing state-dependent transitions in forecasted sensor trajectories.

The rest of this paper is structured as follows: Section \ref{section4.2} presents the detailed methodological framework, elaborating on each component of the proposed architecture, accompanied by necessary mathematical formulations and justifications. Section \ref{section4.3} details the numerical study, encompassing dataset description, data preprocessing procedures, evaluation metrics, comprehensive experimental results, and analyses. Finally, Section \ref{section4.4} concludes this paper by summarizing the key findings.

\section{The Methodology}\label{section4.2}
This section presents a detailed methodology for prognostics under variable operational conditions, built upon a novel multi-headed attention-based fusion neural network (MAFN). The proposed framework explicitly decomposes sensor signals into three essential components: a monotonic degradation trend, distinct operational state embeddings, and residual random fluctuations. The model incorporates advanced deep learning techniques, including attention mechanisms, Bidirectional LSTM networks, and embedding layers, to accurately capture temporal dependencies and operational condition variations.

\subsection{Multi-Headed Attention-Based Fusion Neural Network} \label{section4.2.1}
\begin{figure}[ht]
	\centering
	\includegraphics[scale=0.38]{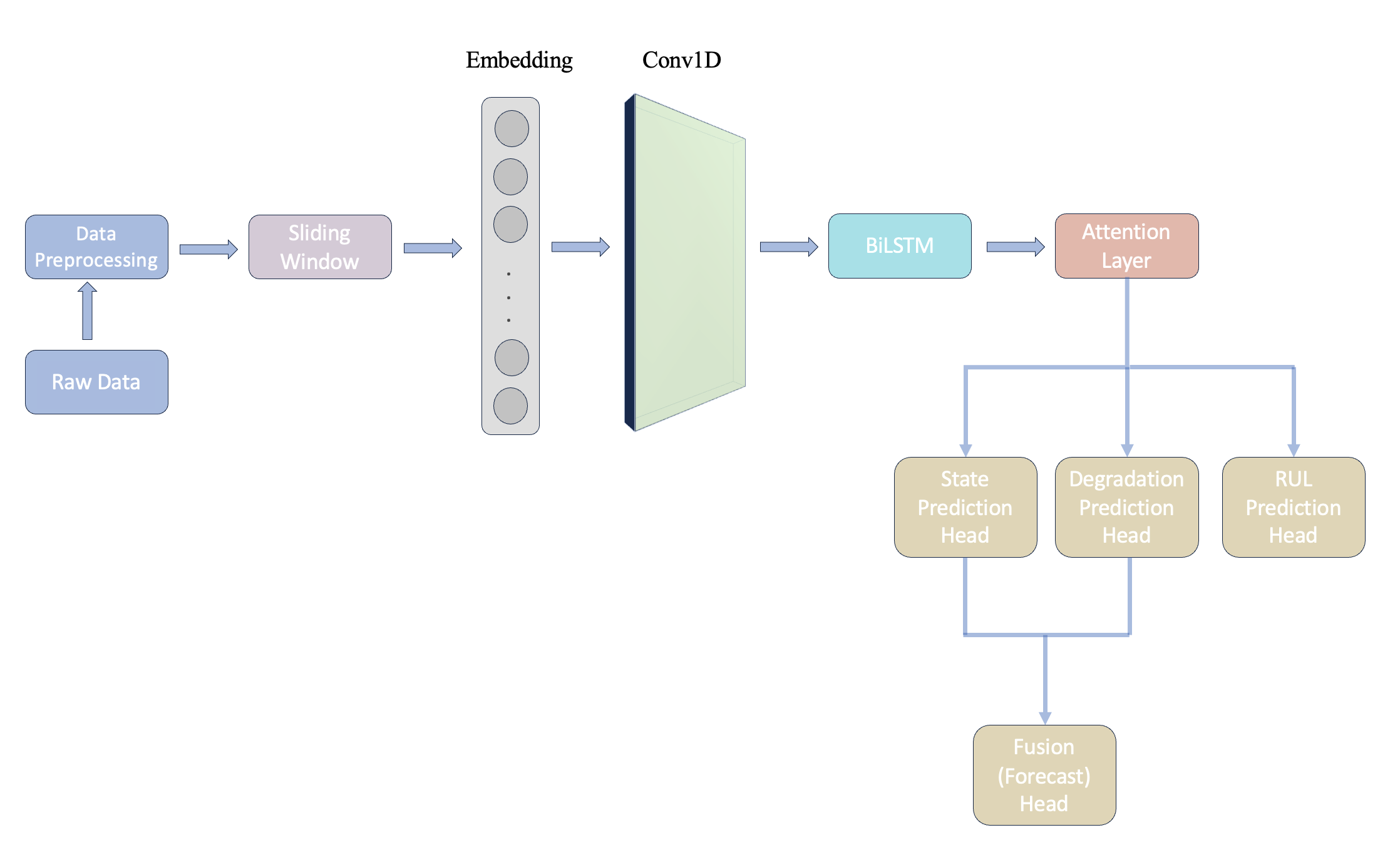}
  \caption{The framework of the proposed network.}
	\label{figure4.p4p14}
\end{figure}

%overall framework
Figure~\ref{figure4.p4p14} summarizes the workflow of the proposed multi‑headed attention‑based fusion network. Starting from raw sensor streams, standard preprocessing is applied and the sequences are then converted into fixed‑length training samples via a sliding‑window module. The windowed inputs proceed through a state‑embedding layer, a 1‑D convolutional layer, and a BiLSTM encoder equipped with an attention mechanism. On top of this shared representation, four task‑specific heads are attached: future state prediction, degradation trend prediction, trajectory forecasting via fusion of trend and state information, and RUL prediction with an asymmetric loss. Beginning with Section~\ref{section4.2.1.1}, we detail these key components one by one.

\subsubsection{Operational Condition Identification and Embedding}
\label{section4.2.1.1}

Operational conditions have a substantial impact on the degradation behavior and consequently on the sensor signals of engineering systems. Therefore, it is critical to explicitly identify and represent these conditions when building accurate prognostic models. In this study, we employ K-Means clustering to systematically classify the operational conditions into distinct categories based on sensor measurements. Specifically, the clustering procedure is applied to the historical sensor data across all relevant operational scenarios, enabling the identification of representative operational states. Mathematically, given a set of sensor measurements \(\mathbf{X} = \{ \mathbf{x}_1, \mathbf{x}_2, \ldots, \mathbf{x}_N \}\), where each \(\mathbf{x}_i \in \mathbb{R}^{D}\) is a \(D\)-dimensional vector corresponding to sensor measurements at time \(i\), the K-Means algorithm partitions the data into \(K\) distinct clusters by solving the following optimization problem:

\begin{equation}
\min_{C} \sum_{k=1}^{K} \sum_{\mathbf{x}_i \in C_k} \|\mathbf{x}_i - \mathbf{\mu}_k\|^2
\end{equation}

where \(C = \{C_1, C_2, \ldots, C_K\}\) represents the set of clusters, \(\mathbf{\mu}_k\) is the centroid of cluster \(C_k\), and \(\|\cdot\|\) denotes the Euclidean norm.

Once the operational conditions are categorized into clusters, we represent these discrete operational states using embedding methods. An embedding layer is introduced to convert categorical operational states into dense vector representations that can effectively capture the relationships and distinctions among states in a continuous feature space. Given an operational state \(s \in \{0, 1, \ldots, K-1\}\), the embedding layer maps \(s\) to a continuous vector representation \(\mathbf{e}_s \in \mathbb{R}^{m}\), defined by:

\begin{equation}
\mathbf{e}_s = \text{Embedding}(s), \quad \text{where } \mathbf{e}_s \in \mathbb{R}^{m}
\end{equation}

Here, \(m\) represents the dimensionality of the embedding space. The learned embedding vectors provide an effective mechanism for capturing the semantic relationships between different operational conditions and significantly enhance the neural network's ability to model condition-dependent degradation behaviors.
%Thus, the combination of K-Means clustering for state identification and embedding for state representation provides a effective foundation for modeling the impact of varying operational conditions on the prognostic outcomes.

\subsubsection{Conv1D Feature Extractor}
\label{section4.2.1.12}

To capture short-range temporal motifs and reduce noise information in multi-sensor signals, we insert a one-dimensional convolutional (Conv1D) layer after state embedding. Let the sliding-windowed, augmented input be
\[
\tilde{\mathbf{X}} = 
\begin{bmatrix}
\tilde{\mathbf{x}}_1^\top\\[-2pt]
\vdots\\[-2pt]
\tilde{\mathbf{x}}_{T_w}^\top
\end{bmatrix}
\in \mathbb{R}^{T_w \times (D+m)}, 
\qquad 
\tilde{\mathbf{x}}_t = [\,\mathbf{x}_t;\,\mathbf{e}_{s_t}\,] \in \mathbb{R}^{D+m},
\]
where \(T_w\) is the window width, \(\mathbf{x}_t\in\mathbb{R}^D\) is the sensor vector at time \(t\), and \(\mathbf{e}_{s_t}\in\mathbb{R}^m\) is the embedding of the operational state at time \(t\) from Section~\ref{section4.2.1.1}. To avoid conflict with the number of clusters \(K\), we denote the convolution kernel length by \(\kappa\).

With \(N_f\) filters, stride \(1\), and \emph{same} padding (so the output length remains \(T_w\)), the \(n\)-th filter is parameterized by a weight tensor \(\mathbf{W}^{(n)}\in\mathbb{R}^{\kappa\times (D+m)}\) and bias \(b^{(n)}\in\mathbb{R}\). The corresponding feature at time \(t\) is
\begin{equation}
z^{(n)}_t \;=\; \phi\!\left(
b^{(n)} \;+\; \sum_{j=0}^{\kappa-1} \big(\mathbf{W}^{(n)}_{j,:}\big)^\top\, \tilde{\mathbf{x}}_{\,t + j - p}
\right), 
\quad t=1,\ldots,T_w,\;\; n=1,\ldots,N_f,
\label{eq:conv1d}
\end{equation}
where $p=\lfloor \frac{\kappa-1}{2}\rfloor$, \(\phi(\cdot)\) is a pointwise nonlinearity (e.g., ReLU), \(\mathbf{W}^{(n)}_{j,:}\in\mathbb{R}^{D+m}\) is the \(j\)-th kernel slice, and indices outside \([1, T_w]\) are handled by zero padding. Stacking filter responses yields
\[
\mathbf{z}_t \;=\; \big[z_t^{(1)},\ldots,z_t^{(N_f)}\big]^\top \in \mathbb{R}^{N_f}, 
\qquad 
\mathbf{Z} \;=\; 
\begin{bmatrix}
\mathbf{z}_1^\top\\[-2pt]
\vdots\\[-2pt]
\mathbf{z}_{T_w}^\top
\end{bmatrix}
\in \mathbb{R}^{T_w\times N_f},
\]
which is then fed to the temporal encoder (BiLSTM + attention in Section~\ref{section4.2.1.3}). Intuitively, the Conv1D layer acts as a learnable bank of short temporal filters that emphasize salient local patterns conditioned on both sensor readings and the embedded operating state, thereby supplying a denoised and information-dense sequence to the recurrent layers.

\subsubsection{Attention-based Bidirectional LSTM Networks}
\label{section4.2.1.2}

Accurate modeling of temporal dependencies and dynamic behavior in sensor data is critical for reliable prognostics, especially in complex engineering systems operating under varying conditions. In this context, Long Short-Term Memory (LSTM) networks have emerged as powerful recurrent neural networks capable of effectively modeling long-term dependencies and handling sequential data. However, standard LSTM architectures often face challenges in identifying and emphasizing the most relevant temporal features from complex sensor signals, which can significantly influence prediction accuracy and interpretability.

\begin{figure}[ht]
	\centering
	\includegraphics[scale=0.38]{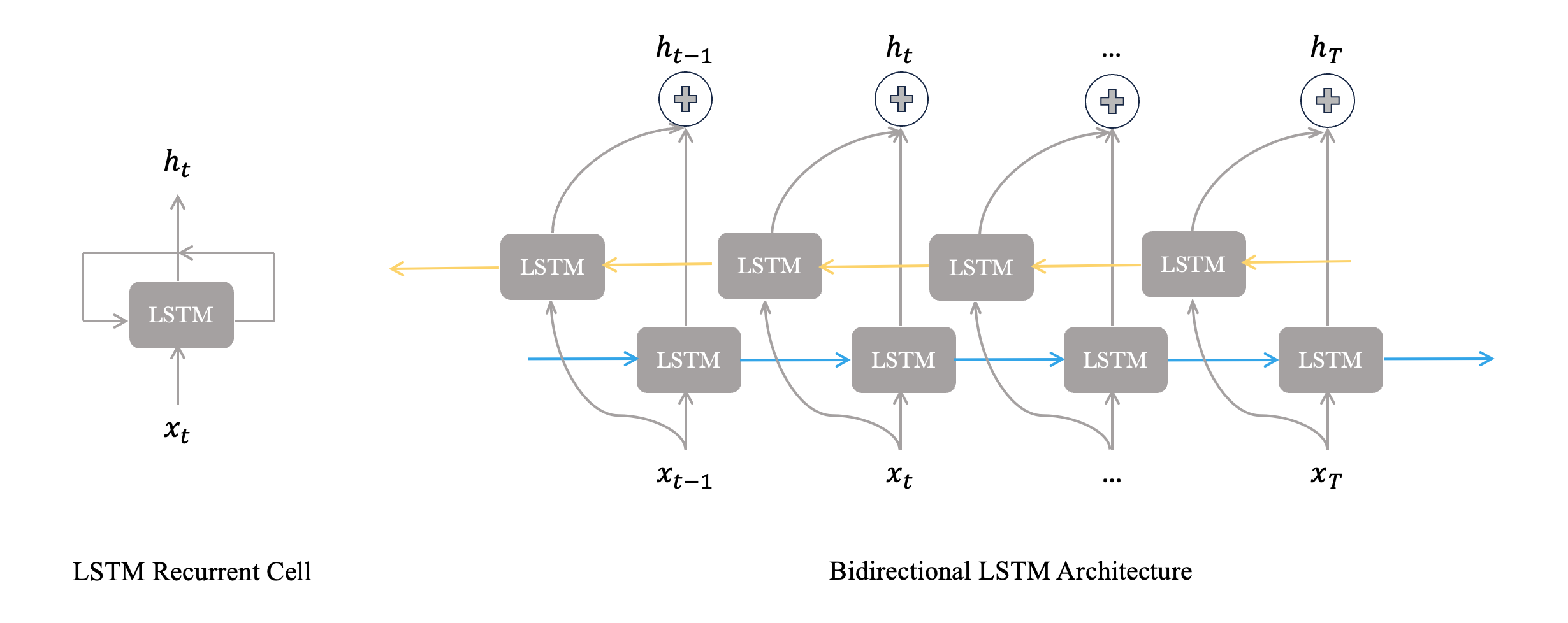}
  \caption{LSTM and BiLSTM architectures.}
	\label{figure4.p4p23}
\end{figure}

To address this limitation, we adopt a Bidirectional LSTM (BiLSTM) network integrated with an attention mechanism \cite{huang2015bidirectional,graves2005framewise}. The Bidirectional LSTM processes input sequences in both forward and backward directions, allowing the network to capture past and future temporal information. Figure \ref{figure4.p4p23} shows the comparison of the LSTM and BiLSTM architectures. Mathematically, the forward and backward passes of the LSTM network at time step \(t\) can be expressed as follows:

\begin{align}
    \overrightarrow{\mathbf{h}}_t &= \text{LSTM}_f(\mathbf{x}_t, \overrightarrow{\mathbf{h}}_{t-1}), \\
    \overleftarrow{\mathbf{h}}_t &= \text{LSTM}_b(\mathbf{x}_t, \overleftarrow{\mathbf{h}}_{t+1}), \\
    \mathbf{h}_t &= [\overrightarrow{\mathbf{h}}_t; \overleftarrow{\mathbf{h}}_t],
\end{align}

where \(\mathbf{x}_t\) represents the input feature vector at time \(t\), \(\overrightarrow{\mathbf{h}}_t\) and \(\overleftarrow{\mathbf{h}}_t\) denote the forward and backward hidden states, respectively, and \([\cdot;\cdot]\) signifies concatenation. This Bidirectional structure captures dependencies from both historical and future perspectives of the time series.

While BiLSTM captures bidirectional temporal context, it does not by itself provide an explicit, data‑dependent weighting across time. In common implementations, the sequence of hidden states is reduced using a fixed rule (e.g., the last hidden state or uniform pooling), which may obscure critical moments that are most indicative of degradation and condition shifts. To address this, we introduce an attention mechanism that learns relevance scores for each hidden state and forms a weighted sum, thereby providing explicit and interpretable emphasis on the most informative time steps for our prediction tasks.

The attention mechanism operates by computing a set of attention weights \(\alpha_t\) for each hidden state \(\mathbf{h}_t\), reflecting the relative importance of each time step. These attention weights are derived through a two-step process: first, computing alignment scores between hidden states and a context vector (often chosen as the hidden state from the last time step or learned as a parameter); second, applying a softmax function to normalize the scores into probabilities:

\begin{align}
    e_t &= \mathbf{v}^\top \text{tanh}(\mathbf{W}_h \mathbf{h}_t + \mathbf{W}_s \mathbf{s}), \\
    \alpha_t &= \frac{\exp(e_t)}{\sum_{j=1}^{T}\exp(e_j)},
\end{align}
where \(\mathbf{h}_t\) is the concatenated hidden state at time step \(t\), \(\mathbf{s}\) represents the summary or query vector, \(\mathbf{v}\), \(\mathbf{W}_h\), and \(\mathbf{W}_s\) are trainable parameters defining the alignment function, and \(T\) is the length of the sequence.

The computed attention weights \(\alpha_t\) subsequently guide the aggregation of hidden states into a fixed-length context vector \(\mathbf{c}\), which summarizes the entire sequence with varying degrees of emphasis placed upon critical time steps:

\begin{equation}
    \mathbf{c} = \sum_{t=1}^{T} \alpha_t \mathbf{h}_t.
\end{equation}
where $\sum_{1}^T\alpha _t = 1$. This context vector \(\mathbf{c}\) embodies the weighted integration of the most relevant historical and contextual information, significantly improving the model's capability to focus on crucial temporal patterns and transitions.

By explicitly incorporating the attention mechanism within the BiLSTM architecture, the proposed model can dynamically capture and emphasize state-dependent shifts, crucial degradation events, and operational condition transitions present in sensor data.

\subsubsection{Degradation Trend Modeling} \label{section4.2.1.3}
Degradation processes represent irreversible and cumulative deterioration phenomena, driven by factors such as wear, fatigue, corrosion, and aging effects. Clearly distinguishing and capturing this monotonic degradation trend is vital because it serves as the core indicator of system health and directly influences the accuracy and reliability of prognostic outcomes.

In real-world applications, sensor measurements typically reflect a mixture of underlying degradation, operational conditions, and random noise. If the degradation component is not explicitly isolated and effectively modeled, prognostic methods risk inaccurately interpreting normal operational fluctuations or random disturbances as actual degradation events, leading to erroneous or misleading RUL predictions.

To address these challenges, the proposed framework explicitly models the degradation trend as a separate component using a dedicated recurrent neural network (RNN) structure designed to capture and represent the progressive nature of degradation processes. Specifically, we employ Long Short-Term Memory (LSTM) networks known for their capability in modeling temporal dependencies. Moreover, to ensure realistic and reliable degradation modeling, our approach introduces monotonicity and smoothness constraints into the training process. The monotonicity constraint explicitly prevents unrealistic reversals or recoveries in the predicted degradation trend, while the smoothness regularization ensures the degradation trend evolves gradually without abrupt, unrealistic fluctuations. These constraints are incorporated via the following penalty terms:

\textbf{Monotonicity constraint}:
\begin{equation}
\mathcal{L}_{\text{mono}} = \sum_{t=1}^{T-1}\max\left(0, -(\hat{y}_{t+1} - \hat{y}_t)\right),
\end{equation}

\textbf{Smoothness constraint}:
\begin{equation}
\mathcal{L}_{\text{smooth}} = \sum_{t=1}^{T-1}(\hat{y}_{t+1}-\hat{y}_t)^2,
\end{equation}

Here, \(\hat{y}_t\) represents the predicted degradation value at time step \(t\). The combined degradation modeling loss is then expressed as:

\begin{equation}
\mathcal{L}_{\text{degradation}} = \mathcal{L}_{\text{mono}} + \lambda_{\text{smooth}} \mathcal{L}_{\text{smooth}},
\end{equation}

with \(\lambda_{\text{smooth}}\) being a hyperparameter to balance monotonicity and smoothness regularization.

\subsubsection{Remaining Useful Life (RUL) Prediction}\label{section4.2.1.4}
Remaining Useful Life (RUL) prediction serves as the core prognostic measure, providing essential information on the time remaining until a system or component reaches its end of life or failure threshold. Accurate RUL predictions are vital for effective maintenance planning, risk management, and decision-making processes, especially within safety-critical engineering systems. The practical implications of inaccurate RUL predictions can be significant, including unexpected equipment failures, increased maintenance costs, and potential safety hazards. Thus, ensuring reliable and accurate RUL estimations is crucial.

In this framework, the RUL prediction is realized through a dedicated neural network head that takes advantage of the comprehensive representation obtained from the shared hidden state and attention-based Bidirectional LSTM networks. Specifically, we employ fully connected (dense) layers to map the shared latent representation into a scalar value representing the predicted RUL.

Formally, the RUL prediction head is mathematically expressed as follows:

Given the shared hidden representation vector \( \mathbf{h} \), the predicted RUL \(\hat{y}_{\text{RUL}}\) is computed through a series of dense layers:
\begin{equation}
\begin{aligned}
    \mathbf{z}_1 &= \text{ReLU}(\mathbf{W}_{z1} \mathbf{h} + \mathbf{b}_{z1}), \\
    \mathbf{z}_2 &= \text{ReLU}(\mathbf{W}_{z2} \mathbf{z}_1 + \mathbf{b}_{z2}), \\
    \hat{y}_{\text{RUL}} &= \mathbf{W}_y \mathbf{z}_2 + b_y,
\end{aligned}
\end{equation}
where \(\mathbf{W}_{z1}\), \(\mathbf{W}_{z2}\), \(\mathbf{W}_y\) and \(\mathbf{b}_{z1}\), \(\mathbf{b}_{z2}\), \(b_y\) represent trainable weight matrices and bias vectors of the dense layers. The rectified linear unit (ReLU) activation function \cite{agarap2018deep} is used to enhance nonlinear modeling capabilities and avoid potential gradient vanishing problems.

Given the practical implications of overly optimistic RUL predictions, especially in safety-critical contexts, an asymmetric loss function is introduced to explicitly penalize late predictions more heavily than early predictions. Late predictions (where the predicted RUL exceeds the actual RUL) pose significant risks as they may lead to unexpected system failures without adequate warning or timely intervention. Such scenarios can result in severe operational disruptions, increased maintenance costs, compromised safety, and potential catastrophic consequences in high-stakes applications. Conversely, early predictions (where the predicted RUL is shorter than the actual RUL) are comparatively conservative, typically leading only to slightly increased maintenance costs without posing significant safety risks.

Therefore, the asymmetric loss explicitly emphasizes penalizing late predictions to ensure that the model provides conservative, safety-oriented prognostic information. The mathematical formulation of the asymmetric loss function is defined as:

\begin{equation}
\mathcal{L}_{\text{RUL}} = \frac{1}{N}\sum_{i=1}^{N} \left[\lambda_{\text{late}}\cdot\text{ReLU}^2(\hat{y}_{\text{RUL}, i} - y_{\text{RUL}, i}) + \lambda_{\text{early}}\cdot\text{ReLU}^2(y_{\text{RUL}, i} - \hat{y}_{\text{RUL}, i})\right],
\end{equation}

where \(\hat{y}_{\text{RUL}, i}\) and \(y_{\text{RUL}, i}\) are the predicted and true RUL values for sample \(i\), respectively, \(N\) is the total number of samples, and \(\lambda_{\text{late}}\) and \(\lambda_{\text{early}}\) are weighting factors controlling the penalty severity for late and early predictions, with the condition \(\lambda_{\text{late}} > \lambda_{\text{early}}\).

\subsection{Fusion Model Construction}\label{section4.2.2}

Section~\ref{section4.2.1} detailed the primary components of our multi-headed deep learning framework. While each component individually captures distinct and critical aspects of system degradation and operational conditions, an effective prognostic model requires the precise integration of these separate streams of information. This subsection focuses on the fusion mechanism designed specifically to combine the outputs of the degradation trend modeling and operational state embedding. Additionally, we introduce tailored loss functions that guide the fusion process to ensure accurate and interpretable predictions of future sensor trajectories and system health states.

\subsubsection{Fusion Mechanism}\label{section4.2.2.1}

The fusion mechanism is a critical component of our multi-headed deep learning framework, responsible for effectively combining distinct information streams-namely, the predicted degradation trends and operational condition embeddings—to generate accurate and interpretable forecasts of future sensor trajectories. The necessity of a fusion mechanism arises from the complex interaction between degradation processes and operational conditions. While degradation represents the intrinsic and irreversible deterioration patterns, operational conditions significantly influence the magnitude and variability of the observed sensor signals. A precise integration of these two distinct sources of information is thus fundamental to capturing realistic degradation behaviors and achieving high prediction accuracy.

Our proposed fusion mechanism explicitly integrates the outputs from the degradation trend model and the state embedding layer. Specifically, the degradation trend model produces a predicted monotonic trend sequence \(\mathbf{D} = \{\mathbf{d}_1, \mathbf{d}_2, \dots, \mathbf{d}_T\}\), representing the underlying deterioration process across a prediction horizon of length \(T\). Concurrently, the state embedding layer transforms discrete operational states into continuous embedding vectors \(\mathbf{E} = \{\mathbf{e}_1, \mathbf{e}_2, \dots, \mathbf{e}_T\}\), effectively encoding operational state characteristics and transitions.

Formally, given the degradation output \(\mathbf{D}_t \in \mathbb{R}^{d_d}\) and state embedding \(\mathbf{E}_t \in \mathbb{R}^{d_e}\) at time step \(t\), we concatenate these two vectors to construct a combined feature vector \(\mathbf{f}_t\):

\begin{equation}
\mathbf{f}_t = \text{concatenate}(\mathbf{D}_t, \mathbf{E}_t), \quad \mathbf{f}_t \in \mathbb{R}^{(d_d + d_e)},
\end{equation}

where \(d_d\) and \(d_e\) are the dimensionalities of the degradation trend and operational state embeddings, respectively. The concatenation explicitly ensures that both degradation dynamics and operational state information are jointly represented.

After concatenation, the combined feature vector \(\mathbf{f}_t\) passes through a carefully designed series of fully connected neural network layers, termed fusion layers, which further learn nonlinear interactions and dependencies between the degradation and state information. These fusion layers are mathematically defined as follows:
\begin{equation}
\begin{aligned}
\mathbf{h}_t^{(1)} &= \text{ReLU}(\mathbf{W}_1 \mathbf{f}_t + \mathbf{b}_1), \\
\mathbf{h}_t^{(2)} &= \text{ReLU}(\mathbf{W}_2 \mathbf{h}_t^{(1)} + \mathbf{b}_2), \\
&\vdots \\
\mathbf{h}_t^{(L)} &= \text{ReLU}(\mathbf{W}_L \mathbf{h}_t^{(L-1)} + \mathbf{b}_L), \\
\hat{\mathbf{y}}_t &= \mathbf{W}_o \mathbf{h}_t^{(L)} + \mathbf{b}_o,
\end{aligned}
\end{equation}
where \(\mathbf{W}_l\) and \(\mathbf{b}_l\) denote trainable weight matrices and biases for the \(l^{th}\) layer, and \(\text{ReLU}(\cdot)\) represents the Rectified Linear Unit activation function. The final predicted forecast \(\hat{\mathbf{y}}_t\in \mathbb{R}^{d_s}\) at time step \(t\) (where \(d_s\) is the number of sensors) explicitly reflects both the underlying degradation trend and the state-dependent operational dynamics.

\subsubsection{Loss Functions and Training Strategy}\label{section4.2.2.2}

%The design and selection of appropriate loss functions are critical to effectively guide the training process and ensure the predictive accuracy, interpretability, and safety of the prognostic model. 
Given the multi-headed structure of our framework, we employ distinct loss functions tailored to each head, capturing their unique characteristics and requirements. Additionally, loss weighting strategies are employed to balance these losses according to their relative importance in the overall predictive task. Specifically, we utilize the following tailored loss functions for each head component:

\paragraph{State Sequence Prediction Loss:} To accurately predict discrete operational condition states, we utilize a masked sparse categorical cross-entropy loss. The mask ensures padding positions (representing unknown future states) do not influence the model training. The loss function is formulated as:
\begin{equation}
\mathcal{L}_{\text{state}} = -\frac{1}{\sum_{t}m_t}\sum_{t=1}^{T} m_t \log p_{t,c_t},
\end{equation}
where \(m_t\) is a binary mask indicating valid (non-padding) positions, and \(p_{t,c_t}\) denotes the predicted probability of the true state \(c_t\) at time step \(t\).

\paragraph{Degradation Trend Loss:} Given that the degradation trend is inherently unknown and monotonic, we employ a loss function with monotonicity and smoothness constraints to depict interpretable degradation accumulation and procedure:

\begin{equation}
\begin{aligned}
\mathcal{L}_{\text{mono}} &= \sum_{t=1}^{T-1}\max\left(0, -(\hat{y}_{t+1} - \hat{y}_t)\right), \\
\mathcal{L}_{\text{smooth}} &= \sum_{t=1}^{T-1}(\hat{y}_{t+1}-\hat{y}_t)^2, \\
\mathcal{L}_{\text{degradation}} &= \frac{\mathcal{L}_{\text{mono}} + \lambda_{\text{smooth}} \mathcal{L}_{\text{smooth}}}{T-1},
\end{aligned}
\end{equation}
where \(\hat{y}_t\) represents the predicted degradation value at time step \(t\), and \(\lambda_{\text{smooth}}\) is a hyperparameter balancing the monotonicity and smoothness constraints.

\paragraph{Forecast Loss:} To effectively capture the behavior of sensor signals and ensure meaningful integration with degradation and state representations, a masked mean squared error (MSE) loss is utilized. This masked loss explicitly ensures that padding positions (representing unknown or future observation points) do not influence the training process. By doing so, the forecast component aids the overall model in representing realistic sensor signal behaviors without necessarily focusing on explicit future sensor values:
\begin{equation}
\mathcal{L}_{\text{forecast}} = \frac{\sum_{t=1}^{T}m_t(\hat{\mathbf{y}}_t - \mathbf{y}_t)^2}{\sum_{t=1}^{T}m_t},
\end{equation}
where \(\hat{\mathbf{y}}_t\) and \(\mathbf{y}_t\) represent the model-generated and actual sensor values at time step \(t\), respectively, and \(m_t\) denotes the corresponding mask for valid positions.

\paragraph{Remaining Useful Life (RUL) Prediction Loss:} Considering the severe operational and safety consequences of overly optimistic (late) predictions, an asymmetric loss function is implemented to explicitly penalize late predictions more heavily than early predictions:
\begin{equation}
\mathcal{L}_{\text{RUL}} = \frac{1}{N}\sum_{i=1}^{N} \left[\lambda_{\text{late}}\cdot\text{ReLU}^2(\hat{y}_{\text{RUL}, i} - y_{\text{RUL}, i}) + \lambda_{\text{early}}\cdot\text{ReLU}^2(y_{\text{RUL}, i} - \hat{y}_{\text{RUL}, i})\right],
\end{equation}
where \(\hat{y}_{\text{RUL}, i}\) and \(y_{\text{RUL}, i}\) represent predicted and true RUL values respectively, and \(\lambda_{\text{late}} > \lambda_{\text{early}}\) explicitly controls penalty severity to ensure conservative predictions.

The overall training loss is computed as a weighted sum of individual losses:
\begin{equation}
\mathcal{L}_{\text{total}} = w_{\text{state}}\mathcal{L}_{\text{state}} + w_{\text{degradation}}\mathcal{L}_{\text{degradation}} + w_{\text{forecast}}\mathcal{L}_{\text{forecast}} + w_{\text{RUL}}\mathcal{L}_{\text{RUL}},
\end{equation}
where \(w_{\text{state}}, w_{\text{degradation}}, w_{\text{forecast}}, w_{\text{RUL}}\) represent hyperparameters balancing each prediction component's contribution to the overall objective. 

\paragraph{Training Strategy:} The proposed model is trained using the Adam optimization algorithm \cite{kingma2014adam}, known for its adaptive learning rate capabilities and efficient handling of sparse gradients. During the training process, we iteratively adjust the loss weights to achieve optimal balance among the prediction tasks, based on the validation performance. Training involves multiple epochs with mini-batch gradient descent, ensuring stable convergence. Additionally, early stopping and model checkpointing techniques are employed to prevent overfitting and retain the best-performing model based on the validation dataset performance metrics.

\section{Numerical Study}\label{section4.3}
In this section, we present a comprehensive numerical study to evaluate the performance of our proposed MAFN network for prognostics under variable operational conditions. This numerical study is divided into two subsections. First, Section~\ref{section4.3.1} provides a detailed description of the dataset and outlines our data preprocessing strategy. Following this, Section~\ref{section4.3.2} presents the experimental results, performance evaluations, and in-depth analyses, highlighting the effectiveness and advantages of our proposed prognostic approach in comparison to state-of-the-art methods.

\subsection{Data Description and Preprocessing}
\label{section4.3.1}
The numerical study in this research utilizes the widely recognized NASA Commercial Modular Aero-Propulsion System Simulation (C-MAPSS) dataset \cite{saxena2008damage}.It has become a benchmark dataset in prognostics and health management research due to its realistic simulation capabilities and complexity of degradation behavior and operational conditions of turbofan engines. The complete C-MAPSS dataset is divided into four distinct sub-datasets: FD001, FD002, FD003, and FD004. These sub-datasets differ primarily in their operational conditions (comprising variations in throttle resolver angle, altitude, and ambient temperature) and the number of fault or failure modes, such as fan and high-pressure compressor (HPC) degradation.

In this study, we specifically utilize the FD002 sub-dataset for model training and evaluation. This sub-dataset consists of degradation data collected from engines operating under six distinct operational conditions, simulating real-world scenarios characterized by varying environmental parameters. Notably, the FD002 dataset features engines experiencing a single fault mode, making it particularly suitable for evaluating models designed to handle operational variability without the complexity of multiple failure types.

The FD002 dataset is structured as follows:

\begin{itemize}
    \item \textbf{Training dataset:} Contains degradation signals from 260 engines, each run to failure under varying operational conditions.
    \item \textbf{Testing dataset:} Contains sensor signals from 259 engines, each truncated at a random time point before reaching their actual failure.
    \item \textbf{RUL dataset:} Provides the actual Remaining Useful Life (RUL) values for each of the 259 testing engines, representing the ground truth used for evaluating prediction performance.
\end{itemize}

Each engine in the dataset is monitored using 21 sensors, which continuously collect data representing the degradation state of the engine throughout its operational lifecycle. However, several of these sensors exhibit signals with negligible variability across the entire operational period, offering minimal degradation-related information. Based on preliminary analyses and existing research, signals from sensors 1, 5, 6, 9, 10, 13, 14, 16, 18, and 19 were found to be relatively constant and non-informative. Thus, these sensors were excluded from subsequent analyses, leaving 11 informative sensors: sensors 2, 3, 4, 7, 8, 11, 12, 15, 17, 20, and 21.

To standardize sensor measurements and facilitate effective learning, we employed min-max normalization independently for each selected sensor across all engines and operational cycles within the training dataset. 
%Specifically, for each sensor \( j \), the normalized sensor reading \( x^{(j)}_{\text{normal}, i}(t) \) from engine \( i \) at time step \( t \) is calculated as:\begin{equation}x^{(j)}_{\text{normal}, i}(t) = \frac{x^{(j)}_{i}(t) - \min\limits_{t,i}(x^{(j)}_{i}(t))}{\max\limits_{t,i}(x^{(j)}_{i}(t)) - \min\limits_{t,i}(x^{(j)}_{i}(t))},\end{equation}where \( x^{(j)}_{i}(t) \) represents the original reading from sensor \( j \) for engine \( i \) at time \( t \). The minimum and maximum values (\(\min\) and \(\max\)) are computed over the entire training dataset, considering all available engines and time steps for each individual sensor.
%Furthermore, the data was structured into sequences using a sliding window method to prepare it appropriately for the temporal modeling capabilities of the proposed neural networks. 
Given the nature of degradation processes, each sequence is associated with a specific RUL target value representing the time remaining until engine failure. As recommended in existing literature, a maximum RUL threshold of 125 cycles was employed, effectively limiting the prediction horizon and mitigating potential issues caused by prolonged non-degradation periods during early engine operation.

\subsection{Results and Analysis}\label{section4.3.2}

In this subsection, we present a detailed evaluation of the proposed MAFN network using two experimental scenarios. We begin by addressing the \textit{first} scenario, in which the original training dataset is partitioned into separate training and testing subsets. The testing subset includes complete degradation trajectories up to actual engine failure points. This setup enables a comprehensive performance evaluation at various cutoff percentages (10\%, 20\%, \dots, 90\%) of engine life cycles.

To systematically assess the predictive performance of the model, we employ the following evaluation metrics:

\begin{itemize}
    \item \textbf{Root Mean Square Error (RMSE)}: Measures the standard deviation of prediction errors, providing insights into overall prediction accuracy:
    \begin{equation}
        \text{RMSE} = \sqrt{\frac{1}{N} \sum_{i=1}^{N}(\hat{y}_{\text{RUL}, i}-y_{\text{RUL}, i})^2}
    \end{equation}
    where \(\hat{y}_{\text{RUL}, i}\) and \(y_{\text{RUL}, i}\) represent predicted and true RUL values for engine \(i\), and \(N\) is the total number of test engines.

    \item \textbf{Relative Error (RE)}: Captures the mean relative deviation of predicted RUL from actual RUL:
    \begin{equation}
        \text{RE} = \frac{1}{N}\sum_{i=1}^{N}\frac{|\hat{y}_{\text{RUL}, i}-y_{\text{RUL}, i}|}{y_{\text{RUL}, i}+\epsilon}
    \end{equation}
    where \(\epsilon\) is a small constant (e.g., \(1 \times 10^{-8}\)) to avoid division by zero.

    \item \textbf{Score}: A custom prognostic metric widely adopted in literature, specifically designed to asymmetrically penalize late predictions more severely than early predictions. It is computed as:
    \begin{equation}
    \text{Score} =\sum_{i=1}^{N} \begin{cases}
        \exp\left(-\frac{\hat{y}_{\text{RUL}, i}-y_{\text{RUL}, i}}{13}\right)-1, & \hat{y}_{\text{RUL}, i} \leq y_{\text{RUL}, i}\\[2ex]
        \exp\left(\frac{\hat{y}_{\text{RUL}, i}-y_{\text{RUL}, i}}{10}\right)-1, & \hat{y}_{\text{RUL}, i} > y_{\text{RUL}, i}
    \end{cases}
    \end{equation}
\end{itemize}

The numerical results at different cutoff percentages are presented clearly in Table~\ref{tab:results_cutoff}.

\begin{table}[h!]
\centering
\caption{Performance results at various cutoff percentages of engine life cycles.}
\label{tab:results_cutoff}
\begin{tabular}{cccc}
\hline
\textbf{Cutoff Percentage} & \textbf{RMSE} & \textbf{RE} & \textbf{Score} \\ 
\hline
10\% & 17.999 & 0.078 & 178.05 \\ 
20\% & 9.822 & 0.045 & 56.84 \\ 
30\% & 7.669 & 0.036 & 38.53 \\ 
40\% & 5.594 & 0.025 & 22.93 \\ 
50\% & 3.311 & 0.018 & 11.65 \\ 
60\% & 2.450 & 0.019 & 9.20 \\ 
70\% & 1.585 & 0.019 & 6.24 \\ 
80\% & 1.237 & 0.022 & 4.37 \\ 
90\% & 2.838 & 0.103 & 11.02 \\ 
\hline
\end{tabular}
\end{table}

The results presented in Table~\ref{tab:results_cutoff} clearly illustrate a notable trend. Initially, as the cutoff percentage increases from 10\% to 80\%, we observe a consistent and substantial improvement across all metrics (RMSE, RE, and Score). This phenomenon occurs because additional historical sensor data provides the model with richer degradation dynamics and patterns, improving prediction accuracy significantly. Specifically, RMSE steadily declines from 17.999 at a 10\% cutoff to a minimum of 1.237 at an 80\% cutoff, indicating increased absolute accuracy in RUL estimation. The RE and Score metrics show similar decreasing trends, emphasizing improved relative accuracy and reduced penalty for prediction errors as more degradation history becomes available.

However, at a 90\% cutoff, a rise in all metrics is observed, where RMSE increases to 2.838, RE rises sharply to 0.103, and Score increases to 11.02. This is attributed to limited remaining data close to actual engine failure, where minor deviations or noise in sensor signals disproportionately affect prediction accuracy. At very high cutoff percentages, prediction errors become more sensitive to minor variations, reflecting the model's difficulty in accurately extrapolating from very short remaining lifespans.

Throughout the simulation, hyperparameter tuning was carefully conducted to optimize the performance of the MAFN network. Parameters such as neuron count in LSTM layers, embedding dimension sizes, and layer configurations underwent systematic evaluation and adjustments based on performance assessed through validation subsets. The training involved multiple epochs and batch size adjustments, ensuring stable model convergence and optimal predictive performance.

In the \textit{second} experimental scenario, we evaluate the proposed MAFN network using the complete FD002 training dataset and subsequently assess its performance on the independent FD002 testing dataset. Unlike the first scenario, which focused on partitioned subsets of the training data, this approach 
allows us to position the performance of our approach within the broader context of existing research, directly assessing its competitiveness and effectiveness compared to established methods. 

The results in Table~\ref{tab:results_compare} demonstrate that, when trained on the complete FD002 training dataset and evaluated on the independent FD002 testing dataset, the proposed MHAF network achieves an RMSE of 13.96 and a Score of 904.88. These values represent the best performance among all compared methods. Specifically, compared to the strongest existing benchmark, RMTF-Transformer \cite{lv2025new}, which reports an RMSE of 14.02 and a Score of 970.84, our method reduces RMSE by approximately 0.43\% and Score by about 6.8\%. This improvement, while modest in RMSE, is more substantial in Score, reflecting the MHAF network’s stronger capability to avoid late predictions, which is a key factor in safety-critical prognostics applications. The observed gains can be attributed to several unique aspects of the MHAF design. First, the decomposition of sensor signals into monotonic degradation trends, discrete operational states, and residual noise enables the network to capture distinct information sources more effectively than models relying solely on sequence-to-sequence architectures. Second, the integration of attention mechanisms within the BiLSTM backbone enhances the model’s focus on informative time steps, particularly under variable operational conditions. Finally, the fusion layer ensures that information from degradation modeling and state embeddings is coherently combined before forecasting and RUL estimation, improving generalization to unseen operational profiles. 

\begin{figure}[ht]
	\centering
	\includegraphics[scale=0.65]{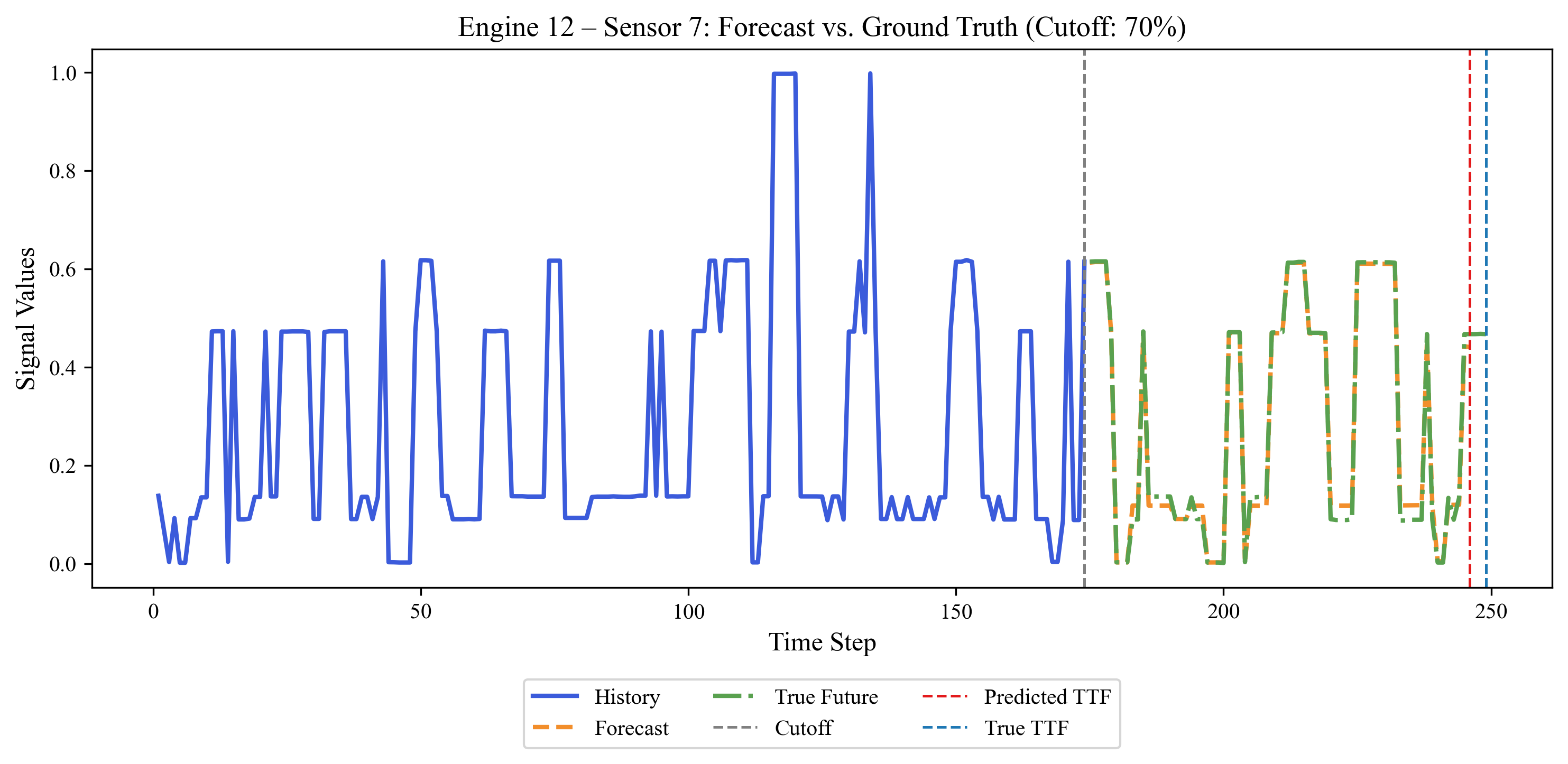}
  \caption{Forecast vs. ground truth signal trajectories (70\% cutoff)}
	\label{figure3.c4p20}
\end{figure}

The figure \ref{figure3.c4p20} contrasts the normalized Sensor \#7 history (blue), the model’s post‐cutoff forecast (orange), and the held‑out ground truth (green) for Engine \#12 with a cutoff at 70\% of the trajectory. Two observations stand out. 
First, the forecast closely tracks the ground truth after the cutoff, reproducing both the local oscillations and the broader level shifts; this indicates that the fusion block successfully conditions the signal evolution on the identified operational state while filtering high‑frequency noise. Moreover, the overall signal pattern appears to be primarily driven by the operational state, whereas within-state fluctuations play a smaller role. Second, the predicted time‑to‑failure (red dashed) is slightly conservative relative to the true TTF (blue dashed)—i.e., it occurs a bit earlier—consistent with the asymmetric loss that penalizes late predictions more heavily. This bias is desirable in PHM practice, where early warnings are safer than late ones.

\begin{table}[h!]
\centering
\caption{Performance comparison of the proposed method and other benchmarks.}
\label{tab:results_compare}
\begin{tabular}{ccccc}
\hline
\textbf{No.} & \textbf{Algorithm} &\textbf{Ref.} & \textbf{RMSE} & \textbf{Score} \\ 
\hline
1 & RUL-RNN& \cite{aggarwal2018two} & 24.67 & N/A \\ 
2 & GCU-Transformer & \cite{mo2021remaining} & 22.81 & N/A \\ 
3 & Semi-Supervised DL+GA & \cite{ellefsen2019remaining} & 22.73 & 3,366 \\ 
4 & BiGRU-AS & \cite{duan2021bigru} & 20.81 & 2,454 \\ 
5 & MT-CNN & \cite{kim2021multitask} & 19.77 & 2,023 \\ 
6 & AEQRNN & \cite{cheng2021autoencoder} & 19.10 & 3,220 \\ 
7 & Siamese & \cite{jang2021siamese} & 18.18 & 1,618 \\ 
8 & MDRNN & \cite{cheng2022multi} & 16.64 & 2,231 \\ 
9 & CBHRL (Regression) & \cite{zhu2024contrastive} & 15.55 & 1,215 \\ 
10 & AMSF-AMoE & \cite{zhou2024adaptive} & 14.42 & 1,077.44 \\ 
11 & RMTF-Transformer & \cite{lv2025new} & 14.02 & 970.84 \\ 
12 & Proposed Network: MHAF & -- & 13.96 & 904.88 \\ 
\hline
\end{tabular}
\end{table}

\section{Conclusions}\label{section4.4}

In this study, we proposed a novel multi-headed attention-based fusion network tailored explicitly for prognostics and health management tasks under variable operational conditions. Our approach effectively addresses critical challenges encountered in accurately predicting the RUL of complex engineering systems, notably the distinct interplay between degradation trends, varying operational states, and random signal fluctuations.

The proposed framework consists of distinct neural network components designed to independently model the underlying monotonic degradation trend, discrete operational state information, and associated noise. By utilizing advanced neural network architectures, including Bidirectional LSTM networks integrated with attention mechanisms, and explicitly embedding operational state information via clustering methods, the model captures rich temporal dependencies and state-dependent signal variations, thereby significantly enhancing predictive accuracy and interpretability. Beyond single‑point RUL estimation, the framework explicitly forecasts future sensor trajectories by fusing the monotonic degradation trend with learned state embeddings, thereby giving planners visibility into the likely evolution of signals. 

Through comprehensive numerical experiments conducted on the NASA C-MAPSS dataset, the model demonstrated superior predictive performance across multiple scenarios. The analysis of results under varying cutoff percentages clearly revealed the model’s increasing accuracy and precision with progressively longer historical sensor data, emphasizing the model’s capability to accurately capture degradation dynamics and operational state transitions. In addition, comparative analysis with recent state-of-the-art prognostic methods published in existing literature highlights that our proposed model not only achieves competitive performance but also exhibits clear improvements in terms of multiple metrics. Such improvements underline the effectiveness and practical value of our proposed framework.

\bibliographystyle{ieeetr}  %order in appearance
\bibliography{M335}

\end{document}